# Divergent strategies and convergent outcomes in autonomous materials discovery

**Jihan Kim**

Department of Chemical and Biomolecular Engineering, Korea Advanced Institute of Science and Technology (KAIST), Daejeon, Republic of Korea. e-mail jihankim@kaist.ac.kr

**Abstract**

Scientific agents are mostly evaluated on whether they complete tasks or recover known results; we instead study variation across repeated open-ended campaigns. Sixteen separately initialized sessions of one model–harness configuration received a frozen database of 12,499 metal-organic frameworks, a methane-storage objective, a pinned protocol and a one-week budget. Strategies diverged into four approaches spanning 100–5,000 screened structures, and eight built 2,253 hypothetical structures. Yet the agents recovered the same materials frontier near 200 $cm^3\ cm^{-3}$, and an independent calculation of the database's porous region found its nine best structures all among their reports. Enforced checks on half the agents raised fresh-run reproduction from one of eight to eight of eight but could not detectably improve conclusion validity, because fifteen of sixteen agents selected the same audit-excluded entry, an incomplete structure whose missing anions created artificial pore volume. Replicated agents thus reveal both robust conclusions and common-mode errors from shared inputs.

With recent advances in large language models (LLMs), scientific agents are beginning to act more like researchers: they can plan chemical experiments[1], use domain software[2], operate autonomous laboratories[3], and predict or generate materials through natural-language interaction[4]. Domain systems can execute end-to-end computational workflows[5], while broader systems can formulate substantial research programs[6,7]. El Agente Q introduced a hierarchical multi-agent architecture that dynamically decomposes natural-language requests, selects tools, and executes and debugs quantum-chemistry calculations[8]. El Agente Quntur builds on this architecture as a research collaborator that reasons over literature and software documentation, designs multistage workflows beyond a single prescribed protocol, and adapts them in research-level case studies[9]. Multi-agent systems have also coordinated substantial biomedical research programs[10]. This shift is especially consequential in computational science, where hypotheses, calculations and analyses all take place in software. An agent can receive an open objective, build its own workflow, spend a finite computational budget and decide what deserves to be reported[11].

Overall, existing benchmarks cover claim re-derivation from code and data, paper replication, algorithm reimplementation, analysis reconstruction, new-data replication and performance-record matching[12−17]. Quntur evaluates 17 quantum-chemistry tasks over five independent runs each and presents open-ended research-level case studies[9]; in computational materials science, AUTOMAT shows that strong coding agents can fail through incomplete procedures, methodological drift and overconfident assessment[18]. Together, these studies establish task-level breadth, adaptability and repeatability. What remains largely unmeasured is the population-level behavior that emerges when replicated agents pursue the same open scientific objective under a shared budget and their trajectories, conclusions and failure modes are analyzed jointly.

This gap matters because repeating the same open mandate changes the question. A population of agents can reveal which scientific behaviors are stable, which vary among agents and whether different strategies recover the same landscape. It can also show how an intervention redistributes effort between exploration and verification. Yet agreement requires careful interpretation[19−21]. Replicas of one model may follow different routes while sharing the same database, software and scientific assumptions. Their convergence may therefore indicate a robust result, a common scientific prior or a common-mode error. Indeed, recent adversarial work shows that a polished autonomous workflow can preserve a misleading scientific premise[22].

To study these behaviors in a controlled scientific setting, we chose methane storage in porous materials. Adsorbed natural gas storage has a clear practical figure of merit: the deliverable capacity between storage and discharge pressures. Standard grand canonical Monte Carlo (GCMC) calculations are inexpensive enough for week-long campaigns, while the search space is large enough to force strategic choices. Decades of experiments and computational screens have mapped the performance landscape unusually well[23−31]: credible leading structures consistently fall near 190–200 $cm^3$ $cm^{-3}$ under standard conditions. Calculations on hypothetical materials with artificially strengthened adsorption sites can reach the low 200s, but these constructs are not experimentally established storage materials[27−29]. Thus, a result near 200 $cm^3$ $cm^{-3}$ is plausible, whereas a value well above it deserves scrutiny; moreover, the optimum of a frozen database under a pinned protocol can be determined independently.

With this case study in hand, we treated sixteen separately initialized sessions of the same model–harness configuration, run under an isolation protocol, as a controlled population of autonomous researchers (Fig. 1). Each began from the same state and received the same task, frozen database, simulation protocol, resources and week-long budget. Eight agents were randomly assigned seven enforceable checks (two capacity thresholds, structural sanity, protocol validity, modification hygiene, finalist reproduction and random auditing) and formed the checked group; the other eight received the same scientific norms without enforcement and formed the unchecked group (Methods; Supplementary Texts S1 and S2). Their claims were scored against an independently calculated reference landscape and a pre-launch integrity audit, both hidden from the agents. The resulting population exposed early strategic divergence, outcome convergence, voluntary uptake of an optional design route, the behavioral effects of enforcement and, ultimately, the evidentiary limits of agreement among replicas.

## Results

### Early strategic divergence from identical starts

All sixteen agents began with the same first step, submitting a simulation job. The median delay was twenty minutes, and every agent submitted one job within three hours. Their paths then separated almost immediately (Fig. 2a), with the first five logged actions forming a unique sequence for every agent. The first declared strategy was logged between launch and hour 34, whereas the first high-accuracy job was logged between hours 1 and 18. The database entry later named the top-performing material ("champion") by fifteen agents was first encountered between eight minutes and 63 hours after launch. Thus, divergence emerged early without assigning the agents distinct research personalities.

This divergence resolved into four broad approaches. Some agents calculated descriptors across the database, trained surrogate models on small samples and simulated model-selected candidates. Others used fast, low-accuracy simulations to maximize coverage. A third group advanced serially through structures on one or two processor cores, whereas a fourth treated the database as a starting point for structural design. Reported screening ranged from 106 to 5,006 structures and from one-tenth of the protocol's floor fidelity to the floor itself; two agents covered most of the database below the floor. Among outputs retained on disk, 925 distinct structures, each counted once across the fleet, were measured at or above the floor and 55 at high accuracy (Supplementary Table S1). Thus, the scientific repertoire was shared, but the routes through it were not.

**Convergence on the same materials frontier**

Despite these different routes, the agents converged on nearly the same set of credible high-performing materials (Fig. 2b,c). After setting aside the audit-excluded entry discussed below, twelve agents named at least one audit-retained material above 180 $cm^3\ cm^{-3}$. Eleven reported the same copper-based pts framework; ten placed it first among retained materials and supplied ten separately generated high-accuracy values between 198.9 and 200.1 $cm^3\ cm^{-3}$. Vanadium-based srs, ytterbium-based nia and indium-based nuc frameworks followed at approximately 197–198, 195–198 and 195–196 $cm^3\ cm^{-3}$. Thus, consensus extended beyond a single material to nearby leading candidates. Ranked-list depth varied because lists were voluntary: nine agents supplied ordered lists, five named lower-ranked materials in prose and two reported no runner-up. Despite uneven reporting and nearly fiftyfold variation in reported search volume, the same leading retained material and nearby competitors recurred.

**Voluntary materials design along an optional route**

The charter named structural modification as a permitted strategy but did not require it, and eight agents chose to pursue it, together constructing 2,253 hypothetical structure files (Fig. 3a). The largest program examined 1,065 interpenetrated MOF parent frameworks, generated 1,713 deinterpenetrated structures by removing individual networks, measured 250 parent–child pairs and identified four unrelaxed candidates with simulated deliverable capacities of 200–204 $cm^3$ $cm^{-3}$. Other agents removed terminal groups or bound water, methylated or fluorinated leading structures, or compressed a unit cell.

The charter required every modified structure to remain charge balanced and to be generated by a documented, reproducible procedure. Five of the eight agents chose operations that preserved valence and charge by construction (Fig. 3a); the remaining three either removed a complete neutral network, preserving stoichiometric ratios and net charge, or scaled the lattice without altering composition. Examples included removing a complete net from an interpenetrated framework (Fig. 3b), replacing a terminal substituent with hydrogen and exchanging aromatic C–H for C–$CH_3$ or C–F (Fig. 3d). Each agent selected starting ("parent") structures according to its own hypothesis about how to exceed its current leader. Intriguingly, a large deinterpenetration program identified a child structure with a capacity of 203.7 $cm^3$ $cm^{-3}$ from a parent that screened at 75.4 $cm^3$ $cm^{-3}$ (Fig. 3b, top). A checked agent applied the same operation to three parents with matched controls and obtained changes of +0.5 $cm^3$ $cm^{-3}$, +9.6 $cm^3$ $cm^{-3}$ and −23.8 $cm^3$ $cm^{-3}$ (Fig. 3b, bottom). Other programs removed terminal groups or water or decorated leading frameworks, generating up to 251 variants (Fig. 3c).

Most programs ultimately exposed the limits of their chosen transformation. Two agents found that their leaders contained nothing to remove. None of one agent's top six candidates carried a removable terminal group, and 399 of another's top 400 carried no removable water. Accordingly, large gains occurred only when the parent capacity was well below the frontier, as in the pair shown in Fig. 3c (from 23.3 $cm^3$ $cm^{-3}$ to 98.1 $cm^3$ $cm^{-3}$). The agent that methylated its leaders found all 20 variants below their parents, including a decrease from 198.3 $cm^3$ $cm^{-3}$ to 187.1 $cm^3$ $cm^{-3}$ for the frontier material in Fig. 3d. It interpreted this result as evidence that methylation moved already highly porous frameworks in the wrong direction. Another fitted a seven-point methylation series whose optimum was the unmodified parent, whereas a third confined its conclusion to the two scaffolds tested. The agent that compressed the champion's lattice by 4% and raised the value to 214 $cm^3$ $cm^{-3}$ described the products as evidence about the protocol's ceiling rather than as materials. In each case, the report preserved the rationale and negative results alongside the measurements.

Program scale differed descriptively between treatment groups. Across the eight agents in each group, five unchecked agents initiated modification programs and together built 2,207 structure files, whereas three checked agents did so and built 46, more often pairing modifications with pristine controls. This count-level contrast was dominated by the largest unchecked program and should not be interpreted as a population-level treatment estimate. Agents in both groups nevertheless reframed screening as a design problem. Three programs adopted the eventual apparent champion as a parent or benchmark, allowing a screening result to become an organizing assumption for downstream work within the same week. Because all modified structures were simulated as constructed under the rigid-framework protocol without geometry relaxation or stability analysis, some transformations may be chemically inaccessible. These structures are therefore evidence of autonomous research behavior rather than proposed materials.

**Redistribution of scientific effort by enforced checks**

The reproduction contrast served primarily as a manipulation check. All eight checked agents independently re-derived their headline numbers from saved inputs, compared with only one of eight unchecked agents (Fisher exact test, $p = 0.001$). Across five measures summarized in Table 1, the checked group performed nineteen verification acts and the unchecked group nine (Mann–Whitney test, p = 0.012); without the fresh-run reproduction that Appendix A mandated, the remaining four measures total eleven against eight. The totals were specified in a pre-analysis amendment filed after the campaign but before any report was read (Methods). The gates therefore induced the verification behavior they explicitly required, particularly fresh-run reproduction.

Exploratory measures numerically favored unchecked agents: they more often modified structures, built cost models, offered a mechanistic account of why their leader led, and attempted to disprove their own claims, recording seventeen behaviors across four measures compared with ten in the checked group. However, this difference did not reach statistical significance (p = 0.14), and the most visible descriptive contrast was dominated by one large unchecked program. These observations suggest, but do not establish, that verification requirements displaced some exploration under a fixed budget.

However, no improvement in conclusion validity was detectable, and with fifteen of sixteen agents naming the same excluded entry the design had essentially no power to detect one. One checked agent and no unchecked agent named an audit-retained structure as champion; this difference was indistinguishable from chance (p = 1.0). That retained answer arose because the agent never examined the excluded entry, not because it recognized the defect. Both groups otherwise recovered the same legitimate frontier and accepted the same excluded deposition. The checks therefore changed behavior but did not target the layer at which the decisive error arose.

**A shared blind spot across divergent routes**

The experiment's strongest convergence provided its weakest independent confirmation. Fifteen of sixteen agents named the same copper sql framework, 2021[Cu][sql]2[ASR]6 (the [FSR]6 file is coordinate-identical), as the best material in the database. The agents reported working capacities of 206.7–207.2 $cm^3\ cm^{-3}$ (Fig. 2b), with a standard deviation of only 0.12 $cm^3\ cm^{-3}$, below the reported simulation uncertainties. This conclusion was shared by seven checked agents and all eight unchecked agents, while the remaining agent never examined the entry: it measured a different deposition sharing the same identifier stem as a cost benchmark and never ran the excluded file at any fidelity.

The value itself was reproducible, but the computation-ready database entry was chemically incomplete. It derives from BSF-71[32], a copper–dipyridyl framework whose square-grid sheets are charge-balanced by non-coordinating $[B_{12}Cl_{12}]^{2-}$ anions held between the layers by weak Cl···H contacts, with acetone on the copper axial sites. The published crystal structure contains the anions and solvent; the database entry retains only $Cu(dpb)_2$, the anions having been removed together with the solvent during curation. The vacated interlayer space appears as pore volume, and the entry's simulated capacity exceeds the leading retained entry by about 7 $cm^3\ cm^{-3}$. Missing counter-ions and charge imbalance are documented curation problems in computation-ready MOF databases[33,35], which sharpens rather than softens the observation: sixteen agents with that literature available treated a known class of artefact as a discovery. One checked agent reproduced the value from archived inputs with a fresh random seed to within 0.05 of its reported simulation uncertainty and passed every enforced audit. Nevertheless, no agent recognized that it had reproduced a property of an incomplete deposition rather than of the reported material.

**Warning signs and their reinterpretation**

At least ten agents recorded signs that this champion material was unusual (Supplementary Table S2), including bare copper centers, low density, very high void fraction, disconnected two-dimensional sheets and extreme surrogate-model residuals; three of the checked agents recorded the bare copper centers under a caveat their appendix required. Each sign was treated as an explanation for exceptional performance rather than a reason to question the structure. One checked agent even built a chemistry audit and correctly calculated a +8 charge across the four copper centers, but then balanced it with negative groups absent from the file, importing expected chemistry into the deposited composition. No agent validated a new chemistry audit against independent known cases.

The agents therefore had evidence of a problem but none successfully applied the charter's charge-balance rule to convert these anomalies into rejection or escalation. The pre-launch audit identified 406 charge-inconsistent entries, six of which also contained enough artificial void space to distort a capacity screen. Varying the search strategy or repeating the calculation cannot correct an error already present in the shared database.

**Position on the independent reference landscape**

The reference calculation measured 2,344 audit-retained structures under the pinned protocol (Methods): 1,497 structures drawn uniformly at random from the database's 9,167 coordinate-distinct audit-retained structures, which give the shape of the landscape, and 847 further structures from a descriptor tail selected by a rule fixed before the calculation, which targets the high-performing frontier. Structures satisfying both were assigned to the random-sample segment, so the two sets are disjoint. The landscape is steep (Fig. 4a). The median deliverable capacity in the random sample is 46 $cm^3\ cm^{-3}$, 82% of the sample lies below 100, 6% lies above 150, and the best sampled structure reaches 196.

Among the 2,344 reference structures measured, which cover the descriptor-selected porous tail and a 16% random sample of the rest, the nine highest point estimates are all materials that at least one agent reported: the copper pts framework at 199.4 ± 1.3, then the vanadium srs, ytterbium nia, aluminium nan, indium nuc, nickel nia, yttrium bcu and two zinc frameworks, down to 190.3. The first structure that no agent named appears at rank ten, at 190.2 (Fig. 4a, inset; Supplementary Table S3). The reference uncertainties overlap within this group, so the ranking is by point estimate rather than a resolved ordering; what is resolved is that the frontier the reference calculation finds is the frontier the agents reported. Eleven agents' best retained claim was the structure with the highest reference point estimate. For the fifteen agent-reported structures measured at floor fidelity, the reference value and the agent's own value differ by −0.2 ± 0.8 $cm^3 cm^{-3}$ on average, and for the nine re-measured at matched claim grade by −0.2 on average and at most 1.9.

Frontier recovery did not scale with search effort (Fig. 4b). Among the twelve agents that stated a screening total, the rank correlation between structures screened and reference top-ten structures reported was 0.10 (Spearman, $p = 0.77$); the four agents without a stated total are shown from their surviving outputs and excluded from the correlation. One agent reported three of the top ten with only 30 surviving simulation outputs, while the agent with the largest surviving footprint, 12,465 structures, reported none. Recovery differed descriptively across the post-hoc strategy classes: descriptor-first agents recovered 2.4 of the top ten on average, breadth-first agents 2.0, serial agents 1.5, and the modification-centered agent none.

In descriptor space, the agents' reported materials occupy one small region of the database, the corner of highest void fraction and largest cavities, and the excluded entry sits inside it (Fig. 4c). The sixteen agents converged by different routes on the same corner of chemical space, and the reference calculation confirms that the corner holds the measured frontier. One frontier member illustrates the limit of any descriptor rule: the aluminium nan framework at rank four was not selected by the porosity rule that defined the reference tail and was measured only because it fell in the random sample, yet one agent had reported it. Because the random sample covers about 16% of the structures outside the tail, a few further frontier members invisible to the descriptor rule may exist unmeasured until the census completes; the statement above is about the measured set.

*Table 1. Comparison of checked and unchecked groups. Unless a total is shown, each value is the number of agents out of eight. Single-row differences use Fisher's exact test, while totals use the Mann-Whitney test.*

| **What was counted** | **Checked (8 agents)** | **Unchecked (8 agents)** | **Difference** |
|---|---|---|---|
| *What they found* | | | |
| Named an audit-retained MOF as the best material | 1 (it never examined the entry) | 0 | not significant |
| Named the audit-excluded entry as the best material | 7 | 8 | not significant |
| Identified the audit-excluded deposition as incomplete | 0 | 0 | not significant |
| Said its own search was finished / unfinished / did not say | 4 / 3 / 1 | 4 / 3 / 1 | not significant |
| *How carefully they checked their own work* | | | |
| Re-ran its final calculation from its own saved files before reporting it | 8 | 1 | p = 0.001 |
| Included job-identifier traceability for reported calculations | 4 | 4 | not significant |
| Corrected one of its own earlier results before the deadline | 6 | 4 | not significant |
| Wrote its own test of whether a structure is physically valid | 1 | 0 | not significant |
| Tested any of its tools against chemistry known from outside the database | 0 | 0 | not significant |
| Total of the five items above | 19 | 9 | p = 0.012 |
| *Exploratory scientific behavior* | | | |
| Ran a test that could have disproved its own claim | 2 | 3 | not significant |
| Offered a mechanistic account of why its leader led | 4 | 5 | not significant |
| Modified structures at all | 3 | 5 | not significant |
| Number of modified structures built, all agents combined | 46 | 2,207 | 48-fold |
| Estimated in advance how much computing its plans would cost | 1 | 4 | not significant |
| Total of the four items above (excluding the count) | 10 | 17 | p = 0.14 |

**Discussion**

Taken together, the sixteen agents revealed a pattern that no single trajectory could have shown. Their strategies diverged almost immediately, with reported search volumes varying by nearly fiftyfold, and several agents used the charter's optional modification route to turn the database-search objective into broader design programs. Despite these differences, the same legitimate materials frontier emerged repeatedly, and the independent reference calculation placed every one of the nine highest reference point estimates among the 2,344 structures it measured on materials the agents had reported (Fig. 4). Distinct scientific paths converged on a stable underlying landscape.

The structure-modification programs illustrate a different consequence of this variation. Several agents treated the stated problem boundary as provisional, turning a database search into studies of deinterpenetration, functionalization, solvent removal or cell compression. The products are hypotheses rather than validated materials, but the behavior is consequential: the charter permitted structural modification but did not require it, and eight agents chose to generate downstream research questions along that route and committed substantial resources to them. Evaluations of autonomous science should therefore consider not only the final answer to the assigned task, but also the research programs that emerge along the way.

The enforced checks predictably increased compliance with required verification: checked agents more consistently reproduced their headline calculations and performed more verification acts. The smaller exploratory total and much larger number of generated structure files in the unchecked group were descriptive, and the aggregate exploration difference was not significant. These patterns therefore suggest a possible budget trade-off but do not establish a general causal effect on exploration. The warning signs surrounding the eventual champion make the more important distinction clear: anomalies prompted several agents to question and rerun the calculation, yet none led them to question the deposited structure itself. Effective checks must test whether the object under study is valid, not merely whether the calculation is reproducible.

The audit-excluded champion qualifies the otherwise encouraging convergence. Neither the model nor the simulator invented or miscalculated the capacity; every agent inherited the same incomplete database object. Replication reduced stochastic uncertainty but not this systematic error. Agreement across repeated agents therefore establishes stability to resampling of one model and workflow, not independent confirmation by different data, methods or scientific perspectives. A random sample of three agents makes the distinction concrete: there was a 98% chance that at least one would report the leading retained material, but certainty that a majority would name the excluded entry (Supplementary Fig. S1). Replication rapidly increased confidence in both the frontier and the shared error.

These results characterize one model–harness, database and task, not a universal agent failure rate. The model is identified in every transcript only by the alias claude-opus-5, so identical weights across the campaign cannot be guaranteed and the study cannot be rerun on a pinned snapshot; the same isolation protocol applied to a second model is the natural next test. Methane storage was chosen because verification is comparatively inexpensive; harder domains may expose additional failure modes. The frozen-investigator design also excluded the adaptive human oversight likely to remain in many deployments. Workspace isolation was specified and audited as a rule rather than enforced through an operating-system boundary, and one shared-directory failure delivered foreign files to two workspaces. Both recipients detected and quarantined the material, no reported value was affected, and excluding both workspaces leaves the main conclusions unchanged. Two further apparatus disturbances touched trajectories rather than results: a fleet-wide infrastructure notice that proved false and was retracted a day later, on which four agents abandoned working grids, and a process reaper in one agent that killed sibling processes on four occasions (Supplementary Section 4.2). Sensitivity analyses excluding the shared-directory source and recipients, or the four agents that acted on the retracted notice, leave the headline counts and both primary contrasts unchanged in direction and significance (Supplementary Table S8); the process-reaper events killed running simulations that were resubmitted, and no reported value depended on a killed process.

Within those limits, the benchmark provides a way to make these distinctions measurable. Its frozen world, known costs, hidden scoring rules and complete trajectories support comparisons across models, charters and validation procedures. The transferable result is not an estimated frequency of any particular behavior, but the need to distinguish three quantities that task-success evaluations can conflate: diversity of research trajectories, stability of conclusions and independence of the scientific substrate. Future studies can test whether communication preserves strategic diversity, whether heterogeneous agents offer stronger confirmation, and which checks detect errors in data rather than in calculations.

## Methods

### Benchmark freezing

The computation-ready CoRE MOF 2024 set[33–35] was assembled once, frozen at N = 12,499 and pinned by a SHA-256 manifest. Source lineage and the full provenance statement are provided in Supplementary Section 1.1. Each workspace verified all 12,499 files on arrival. LC_ALL=C was enforced because the cluster locale otherwise rendered checksum messages in Korean, which caused a naive English-language success check to fail silently. Coordinate-identical entries were scored as one entity, leaving 9,167 distinct structures. Name matching was not used because it both over-merged nonidentical ASR and FSR entries and missed 80 groups of coordinate-identical variants[36,37]. The smoke phase used a 1,731-structure deposition-year subset, while the main campaign used the full frozen set.

**Benchmark integrity audit**

A three-pass charge-accounting procedure was applied before launch. It was validated on 70 ZIF-like structures that returned net charge exactly zero. The procedure flagged 406 compositionally charge-inconsistent entries as a lower bound. Most were too dense to affect a methane-capacity screen; six combined the imbalance with anomalously large void space and were excluded prospectively as capacity artifacts under mechanical rules recorded in the answer key. Ambiguous cases, failure modes and the full exclusion table are given in Supplementary Section 1.2 and Supplementary Table S4. The method is released in the repository, but the agents received neither the flagged list nor the audit tool.

**Protocol pinning and verification**

All agents used RASPA[38] tag v2.0.37, compiled with gcc 4.8.5, with the TraPPE methane model[39], the UFF framework force field[40] and identical simulation settings. The built library was checked rather than trusting the version label alone, and three UFF files were hash-pinned because they contain the settings for truncated and unshifted interactions and disabled tail corrections. Agents could write their own screening, analysis and structure-modification tools, but could not alter the scientific protocol used to support a final claim. A consolidated smoke-phase job reproduced reference values with the exact provisioned toolchain. For the most sensitive excluded structure, the loading at 5.8 bar, N(5.8 bar), in molecules per unit cell was $36.958 \pm 0.600$ compared with $36.841 \pm 0.183$ in the reference calculation. The final working capacity agreed within $0.41\sigma$, and a provisional mid-run value of 206.449 agreed with $207.170 \pm 1.240$ within $0.58\sigma$. Launch required agreement within $3\sigma$. Because RASPA can exit with status 0 after a fatal input error, success checks required the expected output artifacts. All simulations were grand canonical Monte Carlo at 298 K with methane as a single TraPPE united atom and a rigid framework described by UFF with Lorentz–Berthelot mixing, a 12.8 Å cutoff, no tail corrections and no electrostatics; the DDEC6 charges present in the CoRE files were deliberately not read. Deliverable capacity is the absolute loading at 65 bar minus that at 5.8 bar, converted to $cm^3$ of methane at STP per $cm^3$ of framework. The protocol floor was 2,000 initialization and 10,000 production cycles, and claim-grade values used 10,000 and 50,000.

Reference calculation. The exhaustive full-database reference calculation was specified before launch and continues. Before any main-campaign report was read, a pre-analysis amendment filed on 3 September 2026 defined an interim analysis on which Figure 4 rests, in three segments: a uniform random sample of 1,500 coordinate-distinct retained structures (seed 20260903) at floor fidelity; a descriptor tail of the 1,000 retained structures with the highest geometric void fraction plus any with a largest cavity diameter above 15 Å (1,007 structures selected by the rule, 149 of them already in the random sample and credited to it, leaving 858 for the tail), run at floor fidelity in descending order of void fraction, with the top 100 by working capacity to be promoted to claim grade; and every structure named in an agent's final report, re-run at claim grade. Each structure was run once at each grade, so a structure drawn in the sample that also met the tail rule was credited to the sample and the two sets are disjoint. Runs used one core per job. Eleven structures with a single unit cell crashed reproducibly at RASPA start-up on compute nodes only; an output-naming flag verified to leave results bit-identical was applied in an amendment-owned deck tree without touching the sealed inputs (Supplementary Section 5.2). The analysis was frozen on 6 September 2026. The amendment specified a sample of 1,500 and a tail of 858 structures; at the freeze, 3 sampled and 11 tail structures had not completed and are excluded, all were large cells, and where partial running averages were available they lay far below the frontier (Supplementary Section 8.2). The census of all 12,499 structures continues and will be released with the repository. The excluded entry's value under the pinned protocol, 207.17 ± 1.24 $cm^3\ cm^{-3}$, was calculated before launch as part of the answer key.

**Provisioning and blinding**

A script built each workspace from the frozen manifest, created an empty repository and rendered the appropriate charter. The charter asked for the best validated material and a defended judgment about the database ceiling. Only the checked group received an audit appendix. Its seven enforceable requirements included structure sanity, force-field scope, careful handling of modified structures, fresh-run reproduction of the headline value and blinded audits of sampled structures. The other eight agents received the same scientific norms without enforcement. Both groups were instructed to investigate suspicious results and to prefer a correct negative conclusion over an inflated record. Phase tables were filtered to the row belonging to each agent, with no filtering markers left behind. Every completed workspace passed a four-class disclosure scan, and session-loading paths were checked for instructions outside the experiment. The detected disclosure classes and the scanner design are described in Supplementary Section 3.

The investigator made no scientific decisions during the campaign. A separate controller provisioned workspaces, enforced budgets, collected records and handled escalations under a pre-registered clause that guaranteed logging and reading but not an answer. Fifty-eight escalations were received, 54 about infrastructure and 4 about the charter; 49 were answered with operational information such as scripted repairs, budget rulings and environment facts, 42 of them before the recipient's campaign closed, and no scientific guidance was given (Supplementary Section 4.2). Charter questions received the instruction to proceed under the charter and log the ambiguity. The agents were not told that other agents or treatment groups existed.

**Agent runtime**

Each of the sixteen agents was a separate session of the same frontier model, recorded in every transcript as claude-opus-5 with no finer snapshot identifier, accessed through Claude Code 2.1.233 between 29 August and 1 September 2026 (UTC) and run on a dedicated host with one transcript directory per agent. Restart after process death was capped at three attempts, and each restart was logged. Liveness was determined from transcript growth in bytes because the smoke phase showed that heartbeat files and modification times could indicate activity for blocked or dead sessions. Main sessions ran headlessly, unlike the interactive smoke phase. A billing preflight checked account headroom before launch. One agent (rep01) served as an instrumented launch gate and had to pass nine assertions, including transcript growth, billing status, manifest verification, successful job round trips, nonzero meters, watchdog activity and a writable escalation channel. The remaining fifteen agents launched in two waves after the gate passed. Workspaces were not reset. Deadlines were extended uniformly by 4.5 hours for a fleet pause during a host migration, and, in addition to that, for eleven agents by restoration of time lost to controller faults under a standing rule: 14.0 to 15.6 hours for ten agents affected by one fault batch, 0.2 hours for one other agent, and a further 9.6 hours for one of the ten, whose restorations therefore totalled 24.7 hours (Supplementary Table S5). No agent reached its deadline: twelve ended at the spend limit and four by filing their final report, and active sessions lasted 33 to 79 hours of the 168 allowed. General numerical Python libraries were available, but materials-specific packages such as Zeo++, pymatgen and MOFid were absent and could not be acquired through the network allowlist. All agents used the same cluster account. Isolation was enforced by charter and audited through scheduler records and transcripts, not by an operating-system boundary. One confirmed shared-directory failure affected two workspaces: rep01 received

rep09's draft report, and rep03 had its state file overwritten by rep09's. Both detected and quarantined the foreign material, no reported value was affected, and the report recipient later withheld one analysis it could not show was its own (Supplementary Section 4.2).

**Budgets and metering**

Each agent received 1,610 CPU-hours, compared with an estimated 22,873 CPU-hours for naive exhaustive screening. It also received 32 million tokens and a US$280 spending limit. The spending limit was necessary because cache reads accounted for 59% of the smoke-phase cost and greatly exceeded the billable token count, and it was the binding constraint for twelve of the sixteen agents. Spending was calculated from session transcripts at published rates, including cache reads, and was enforced with warnings at 75% and a hard stop at 100%. Compute was calculated from completed job records after the smoke phase showed that scheduler-only accounting under-reported use by more than twofold. Concurrency was capped at 12 queued jobs per agent and 192 across the fleet. Supplementary Section 4.3 gives the meter validation and enforcement rules.

**Pre-registration and analysis**

Before launch, the independent full-database calculation was specified and scheduled for the scoring phase. The hidden answer key combined that reference landscape with the literature-calibrated ceiling and the pre-launch integrity audit. It used a band rather than an exact ordering when coordinate-identical depositions differed within their observed provenance spread. This made the final conclusions scorable without revealing a target to the agents. Every scored outcome was anchored to the hidden key or extracted from auditable records.

The two charter versions, seeded group assignment, budgets, scoring rubric, analysis plan and answer-key census were sealed by a public-hash commit before launch. A budget amendment made before any agent launched reduced the planned campaign from 20 to 16 agents and from 10 to 7 days. Five post-seal amendments were logged. Rev 21 clarified checked-group audit procedures; Revs 22 to 25 concerned pinned-file scope, cost and context discipline and endgame conduct. Revs 21 to 23 preceded the wave launches but reached the already-running rep01 by notice; Revs 24 and 25 were delivered to agents still active. None changed the scientific task, simulation protocol or answer key (Supplementary Section 2.3). The final 16 were selected mechanically as the first eight agents in each group from the sealed assignment. Primary proportions carry Wilson 95% intervals: 15 of 16 agents named the excluded entry as champion (0.72 to 0.99), 11 of 16 reported the leading retained material (0.44 to 0.86), and 8 of 8 checked against 1 of 8 unchecked agents reproduced their headline value (0.68 to 1.00 against 0.02 to 0.47). Excluding both workspaces that received foreign files, these become 13 of 14, 10 of 14 and 7 of 7 against 1 of 7; the reproduction contrast remains significant (two-sided Fisher exact $p = 0.005$), and the champion contrast remains nonsignificant ($p = 1.0$). Group differences use Fisher's exact test for individual behaviors and the Mann-Whitney test for behavior totals. A conditional second wave of four deferred agents was specified if the primary contrast, the proportion of agents naming an audit-retained entry as champion, returned a p value between 0.05 and 0.25, subject to a pre-registered account-limit condition. The primary contrast returned $p = 1.0$, outside that window, so the second wave was not triggered and the account-limit condition was never tested. The trigger did not apply to the exploratory behavior totals. The controller independently reran every headline value from archived inputs before scoring. Scored conduct measures were extracted from the final reports through mechanical text rules (Supplementary Table S6). A logged action is one entry in an agent's own append-only campaign log. The four strategy classes were assigned post hoc by the investigator from each agent's final report, using the agent's own description of its search order, without a second coder; they are descriptive labels and enter no statistical test. Because one branch of the reproduction rule coincides with Appendix A's wording, the two checked agents matching only that branch were verified directly against their audit records, which show a second claim-grade run of the champion with a different seed before filing. Manual overrides in the extraction only ever withheld credit that a rule would have granted. A pre-analysis amendment filed before any report was read replaced planned hand scoring with these objective extractions.

**Smoke phase**

Two agents, one from each group, tested the apparatus for 65.5 hours and were excluded from analysis by pre-registration. They exposed more than a dozen instances of one recurring defect class in which a success signal referred to the wrong object. Examples included a simulation that exited normally after a fatal input error, a watchdog that had not been tested on a live process, an unexecuted job-submission module and a provisioner that would have supplied the wrong database while reporting successful verification. These failures led to a standing rule that success must be judged from the produced artifact. One smoke agent also lost 39.2 hours to an unanswered account-limit dialog, which motivated headless execution and the billing preflight. Supplementary Section 5 contains the full defect ledger and charter revision trail.

**Agent disclosure and investigator interaction**

The agents were not informed that they were part of a replicated evaluation. The complete instructions are released verbatim. The investigator's only contacts with agents were escalation answers delivered and timed by the controller under the pre-registered clause. Operational interventions outside that clause, including one retracted infrastructure notice, are listed in Supplementary Table S5.

**Data availability**

The repository at https://github.com/jihankim929/replicate-study contains the benchmark

manifest, both charter renderings, the controller and integrity-audit code, the scoring rubric, the answer key, the smoke-phase records, the analysis tables and scripts behind Figures 2 to 4 and Supplementary Fig. S1, the descriptor table, and all sixteen agent repositories with complete git histories, audit ledgers and verbatim session transcripts. The pre-registration hash has been public since launch, and the submitted version is archived at https://doi.org/10.5281/zenodo.22520942. Structure files are distributed as a manifest of identifiers and hashes rather than as coordinates, because part of the frozen set derives from the Cambridge Structural Database. The RASPA input decks and output files supporting every headline value, that is the 55 structures the agents measured at high accuracy and the reference calculation's top 100, are deposited with the archived release; the remaining raw simulation output (several gigabytes of RASPA output across the fleet) is available from the author on request, and the reference results are released as a table of values per structure and setting.

**Code availability**

Custom code for benchmark provisioning, budget control, integrity auditing, behavioral extraction and analysis is available at https://github.com/jihankim929/replicate-study and is archived with the repository release at https://doi.org/10.5281/zenodo.22520942.

**Use of generative artificial intelligence**

Beyond the sixteen evaluated sessions, Claude Code and ChatGPT were used under the author's direction to support code development, analysis planning, and manuscript drafting and editing. The author retained responsibility for the study design, scientific judgments, verification of the reported results and the integrity of the final manuscript.

**Acknowledgements**

This work was supported by the National Research Foundation of Korea (NRF) (RS-2024-00451160 and RS-2024-00435493).

**Author contributions**

J.K. conceived and designed the study, developed and supervised the computational infrastructure, analyzed and interpreted the data, and wrote the manuscript.

**Competing interests**

The author declares no competing interests.

## Figure legends

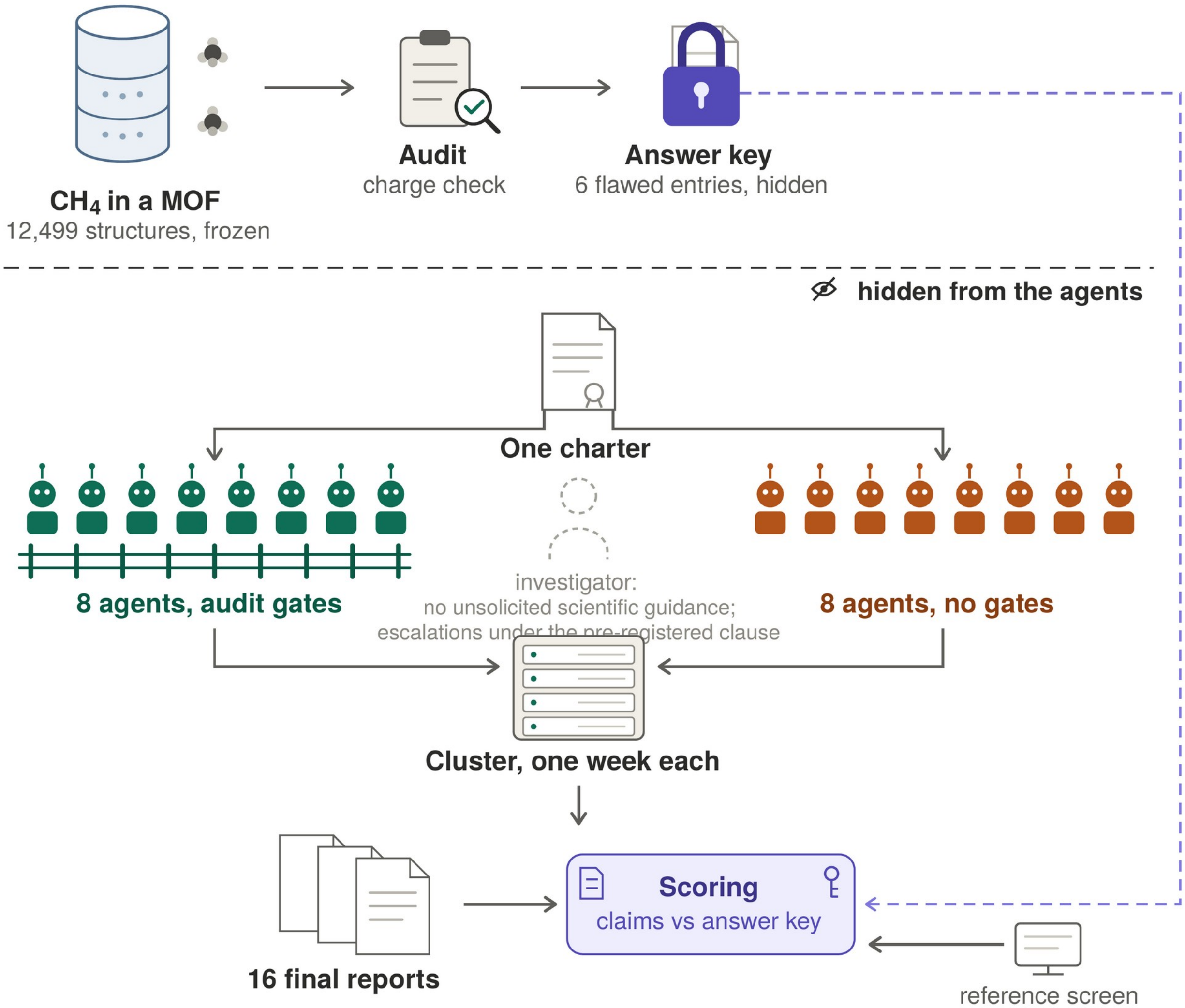

**Figure 1. Design of the experiment.** A frozen database of 12,499 computation-ready metal-organic frameworks, a pre-launch integrity audit and the resulting answer key were fixed before launch and hidden from the agents. Sixteen separately initialized sessions of the same model–harness configuration, run under an isolation protocol, received the same charter, database, simulation protocol and budget; eight also received an appendix of seven enforceable checks (checked group) and eight did not (unchecked group). The investigator had no scientific contact with any agent after launch. Each agent submitted its own simulations to a shared cluster and filed a final report, which was scored against the answer key and the independent reference calculation.

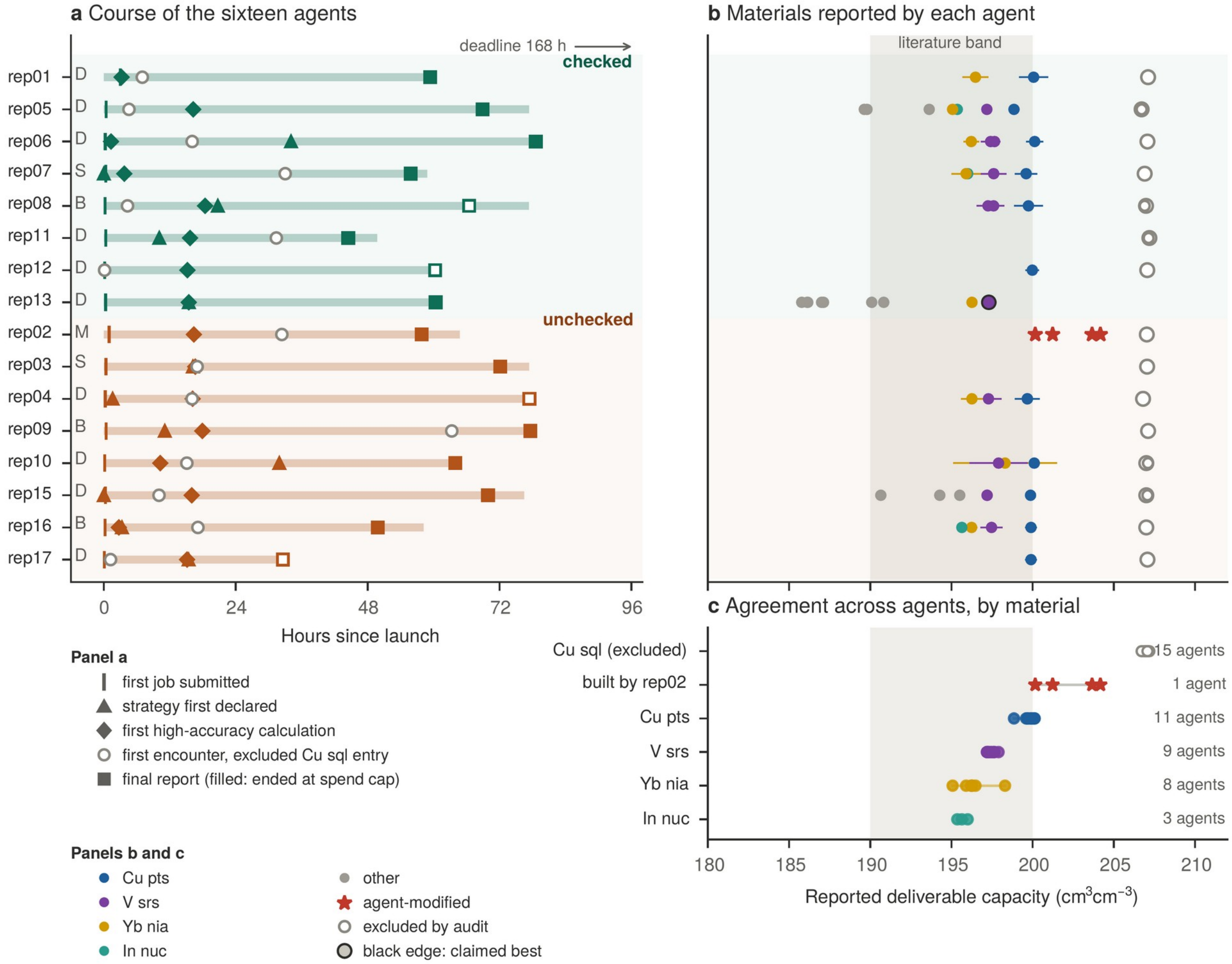

**Figure 2. Divergent courses and convergent claims across sixteen agents. Rows are the sixteen agents, checked (green) above unchecked (orange), the same agent in each row of a and b; letters give the search strategy (D, descriptor-first; B, breadth-first; S, serial; M, modification-centered). a, Course of each campaign in hours since launch. Bars span the active session; markers show the first job submitted, the first declared strategy, the first log entry for a high-accuracy calculation (at least 10,000 initialization and 50,000 production cycles), the first encounter with the audit-excluded Cu sql entry, and the final report (filled, ended at the spend limit; open, ended by filing). Times are the first log entries matching each pattern, not verified completions; the encounter marker is the first contact with either file of the excluded entry (rep13 had none), and rep08's is known only to the day and plotted at its start. No agent reached the 168-hour deadline. b, Deliverable capacities in each final report: the claimed best material has a black edge, retained entries are colored by material, audit-excluded entries are open grey, and agent-built structures are red stars, with the agents' own stated uncertainties, which under the pinned protocol are RASPA block-averaged standard deviations over five blocks; all error bars use that convention. The band marks the literature range of credible capacities, 190 to 200 $cm^3\ cm^{-3}$. Passing mentions below 180 $cm^3\ cm^{-3}$ and values labeled as absolute uptake are omitted. c, The same values grouped by material, with the number of agents reporting each. Full identifiers and every reported structure are in Supplementary Table S7.**

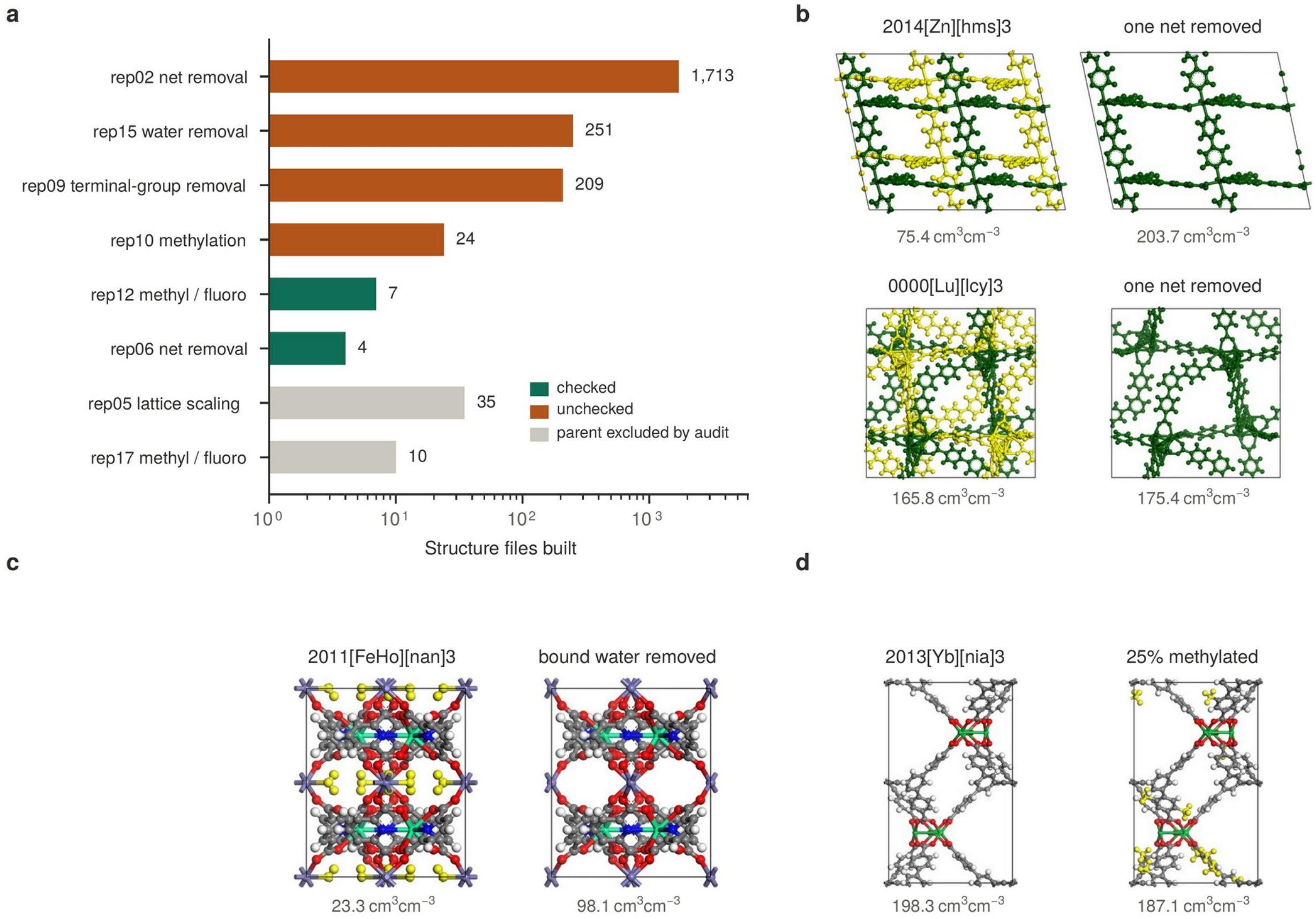

**Figure 3. Voluntary materials design along a permitted but unrequired route. a, Structure files built by the eight agents that modified database entries, colored by group; grey bars are programs whose parent was the audit-excluded entry. b to d, A parent (left) and the agent-built child (right) for three transformations, with the deliverable capacity reported by the agent beneath each structure. Examples follow a fixed rule: one per transformation, a parent retained by the audit, and the pair for which the agent reported both values; where two agents made the same transformation, one from each group is shown. b, Removal of one network from two-fold interpenetrated frameworks (networks in green and yellow), by an unchecked agent (2014[Zn][hms]3[ASR]1, parent at screening fidelity) and a checked agent (0000[Lu][lcy]3[ASR]1, both values at one setting). c, Removal of twelve metal-bound water molecules (yellow) from 2011[FeHo][nan]3[FSR]1. d, Methylation of 25% of aromatic C-H sites (added atoms in yellow) on 2013[Yb][nia]3[ASR]1, a frontier material of Fig. 2. All children were simulated as built, without geometry relaxation or stability assessment.**

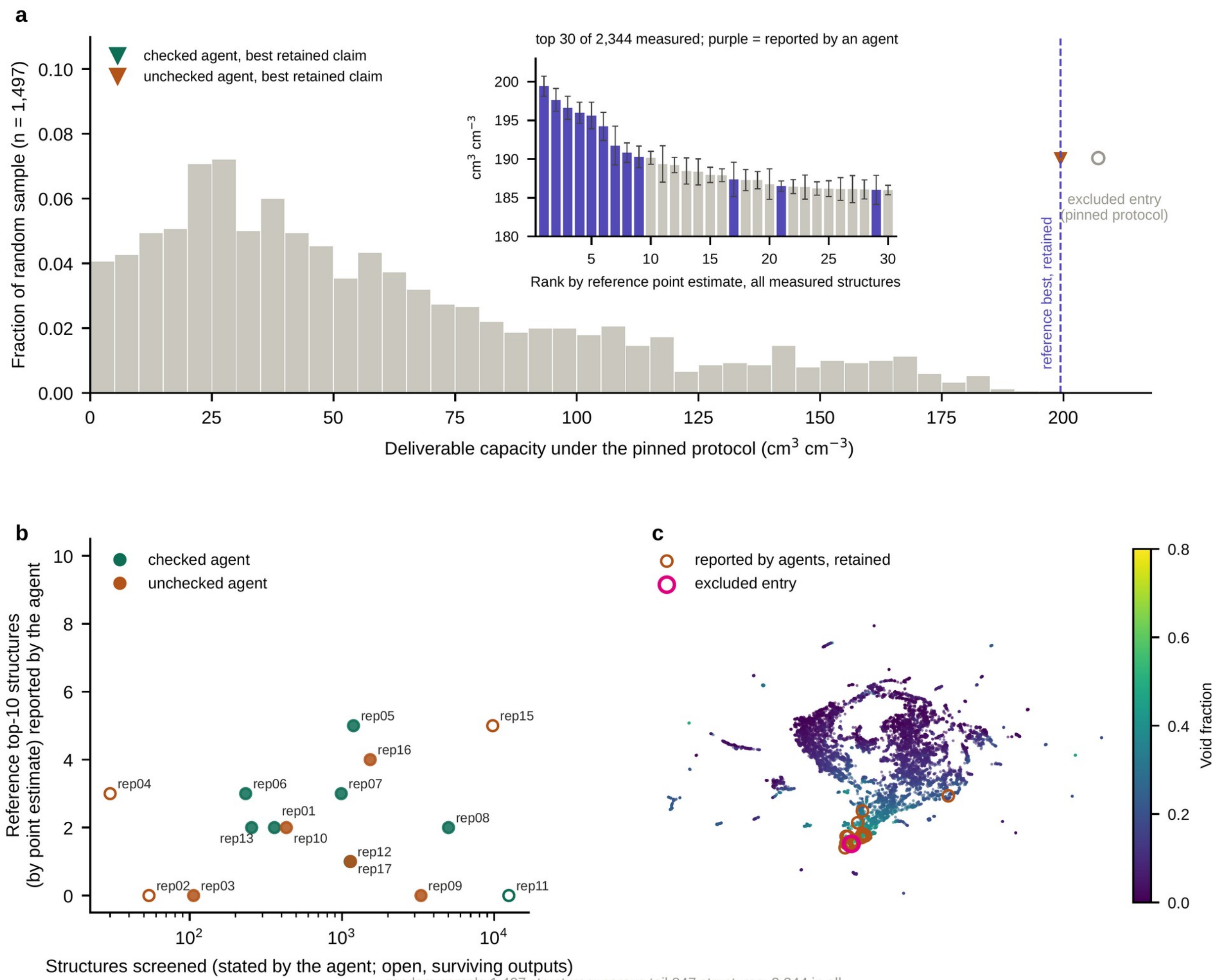

**Figure 4. The agents' claims on the independent reference landscape. a, Distribution of deliverable capacity under the pinned protocol for 1,497 audit-retained structures drawn uniformly at random, as a fraction of the sample. Triangles mark the reference value of each agent's best retained claim (green, checked; orange, unchecked); the dashed line is the highest retained reference point estimate; the open circle is the audit-excluded entry at its pre-launch value under the same protocol, 207.17 ± 1.24 $cm^3$ $cm^{-3}$, which the reference ranking of retained structures does not include. Inset, the thirty highest reference point estimates among all 2,344 structures measured (the random sample plus 847 descriptor-tail structures, disjoint sets), with reference uncertainties; purple bars are structures reported by at least one agent. b, Number of the ten highest reference point estimates that each agent reported, against the number of structures the agent stated it screened (filled) or, for the four agents that stated no total, the number of distinct structures with surviving simulation output (open). c, UMAP embedding of all 12,499 database entries from composition and geometric descriptors (Supplementary Section 8.4), colored by geometric void fraction on the audit's scale; orange circles are structures reported by the agents, magenta the excluded entry.**